%% file: main.tex
\documentclass[conference]{IEEEtran}
\IEEEoverridecommandlockouts

\RequirePackage[absolute]{textpos}
\DeclareRobustCommand*{\copyrightnote}{%
  \begin{textblock}{85}(17.5,256.75)
      \scriptsize{\noindent \copyright 2021 IEEE. Personal use of this material is permitted. Permission from IEEE must be obtained for all other uses, in any current or future media, including reprinting/republishing this material for advertising or promotional purposes, creating new collective works, for resale or redistribution to servers or lists, or reuse of any copyrighted
component of this work in other works.}%
  \end{textblock}%
    }

\usepackage[giveninits=true,sorting=none,style=ieee]{biblatex}
\AtEveryBibitem{\clearfield{pages}} 

\usepackage{placeins}

\usepackage{cancel}
\include{commands}
\usepackage{amsmath,amssymb,amsfonts, bbm}
\usepackage{algorithmic}
\usepackage{graphicx}
\usepackage{subfigure}
\usepackage{textcomp}
\usepackage{xcolor}
\usepackage{hyperref}
\def\BibTeX{{\rm B\kern-.05em{\sc i\kern-.025em b}\kern-.08em
    T\kern-.1667em\lower.7ex\hbox{E}\kern-.125emX}}

\usepackage[capitalize]{cleveref}
\crefname{figure}{Fig.}{Figs.}
\crefname{table}{Table}{Tables}
\crefname{equation}{}{}
\Crefname{figure}{Fig.}{Figs.}
\Crefname{table}{Table}{Tables}
\Crefname{equation}{Equation}{Equations}

\newcommand{\norm}[1]{\left\lVert#1\right\rVert}

\begin{document}

\title{
Learning through structure: towards deep neuromorphic knowledge graph embeddings
\\
\thanks{
\eqcontr These authors contributed equally to this work.}
}

\author{\IEEEauthorblockN{Victor Caceres Chian\eqcontr, Marcel Hildebrandt\eqcontr, Thomas Runkler, Dominik Dold\eqcontr}
\IEEEauthorblockA{
\textit{Siemens AG Technology}\\
80331 Munich, Germany 
}}

\maketitle
\copyrightnote
\thispagestyle{plain}
\pagestyle{plain}
\begin{abstract}
\input{content/abstract}

\end{abstract}


\section{Introduction}
\input{content/introduction} 

\section{Background and Notation}
\label{sec:backround_and_notation}
\input{content/background}

\section{Frozen graph convolutions}
\label{sec:frozen_graph_convolutions}

\input{content/FRGCN}

\section{Spike-based graph embedding}
\label{sec:spike-based_graph_embedding}
\input{content/SRGCN}


\input{content/conclusion}

\section*{Acknowledgment}
\input{content/acknowledge}

\printbibliography
\addcontentsline{toc}{section}{References}
\section*{Supplementary Material: Simulation details}
\input{content/details}

\end{document}

%% file: commands.tex
\newcommand{\ent}[1]{\pmb{e}_{#1}}
\newcommand{\stime}[1]{\pmb{t}_{#1}}

\newcommand{\entT}[1]{\pmb{e}^\intercal_{#1}}

\newcommand{\vrel}[1]{\pmb{r}_{#1}}

\newcommand{\lone}[1]{\| #1 \|}

\newcommand{\heaviside}[1]{\Theta\left( #1 \right)}
\newcommand{\taus}{\tau_\mathrm{s}}

\newcommand{\thresh}{u_\mathrm{th}}

\newcommand{\triple}[3]{($#1$,\ \nolinebreak $#2$,\ \nolinebreak $#3$)}

\newcommand{\eqcontr}{$^*$}
\newcommand{\tcol}[1]{\textcolor{gray}{#1}}
\newcommand{\tcolt}[1]{\textcolor{gray}{#1}}
\newcommand{\best}[1]{\textbf{#1}}

%% file: content/abstract.tex
Computing latent representations for graph-structured data is an ubiquitous learning task in many industrial and academic applications ranging from molecule synthetization to social network analysis and recommender systems. 
Knowledge graphs are among the most popular and widely used data representations related to the Semantic Web. Next to structuring factual knowledge in a machine-readable format, knowledge graphs serve as the backbone of many artificial intelligence applications and allow the ingestion of context information into various learning algorithms.
Graph neural networks attempt to encode graph structures in low-dimensional vector spaces via a message passing heuristic between neighboring nodes.
Over the recent years, a multitude of different graph neural network architectures demonstrated ground-breaking performances in many learning tasks.
In this work, we propose a strategy to map deep graph learning architectures for knowledge graph reasoning to neuromorphic architectures. Based on the insight that randomly initialized and untrained (i.e., frozen) graph neural networks are able to preserve local graph structures, we compose a frozen neural network with shallow knowledge graph embedding models. We experimentally show that already on conventional computing hardware, this leads to a significant speedup and memory reduction while maintaining a competitive performance level. Moreover, we extend the frozen architecture to spiking neural networks, introducing a novel, event-based and highly sparse knowledge graph embedding algorithm that is suitable for implementation in neuromorphic hardware.

%% file: content/introduction.tex
A quintessential aspect of neural networks is the adjustment of their synaptic weights during training to optimize a given cost function \cite{lecun2015deep, rumelhart1988learning}.
Since the cost function is, in principle, arbitrary, neural networks have emerged as flexible models for a multitude of applications \cite{krizhevsky2012imagenet, silver2017mastering, brown2020language}.
However, even with static weights, neural circuits have been shown to possess intriguing properties, e.g., for information processing \cite{jaeger2001echo,natschlager2002liquid,saxe2011random}, to support learning \cite{sacramento2018dendritic,guerguiev2017towards,raman2021frozen}, for transfer learning \cite{yosinski2014transferable} or to enable efficient hardware realizations of neural networks with reduced silicon area and power consumption \cite{isikdogan2020semifreddonets}.
We propose a frozen architecture inspired by relational graph convolutional networks (R-GCN) \cite{schlichtkrull2018modeling} that is suitable for efficient hardware realizations like neuromorphic systems \cite{thakur2018large} while offering advanced functionality for reasoning on symbolic data like knowledge graphs (KGs) -- despite having random and static weights. 

\begin{figure}[t]
\centering
\includegraphics[width = \columnwidth]{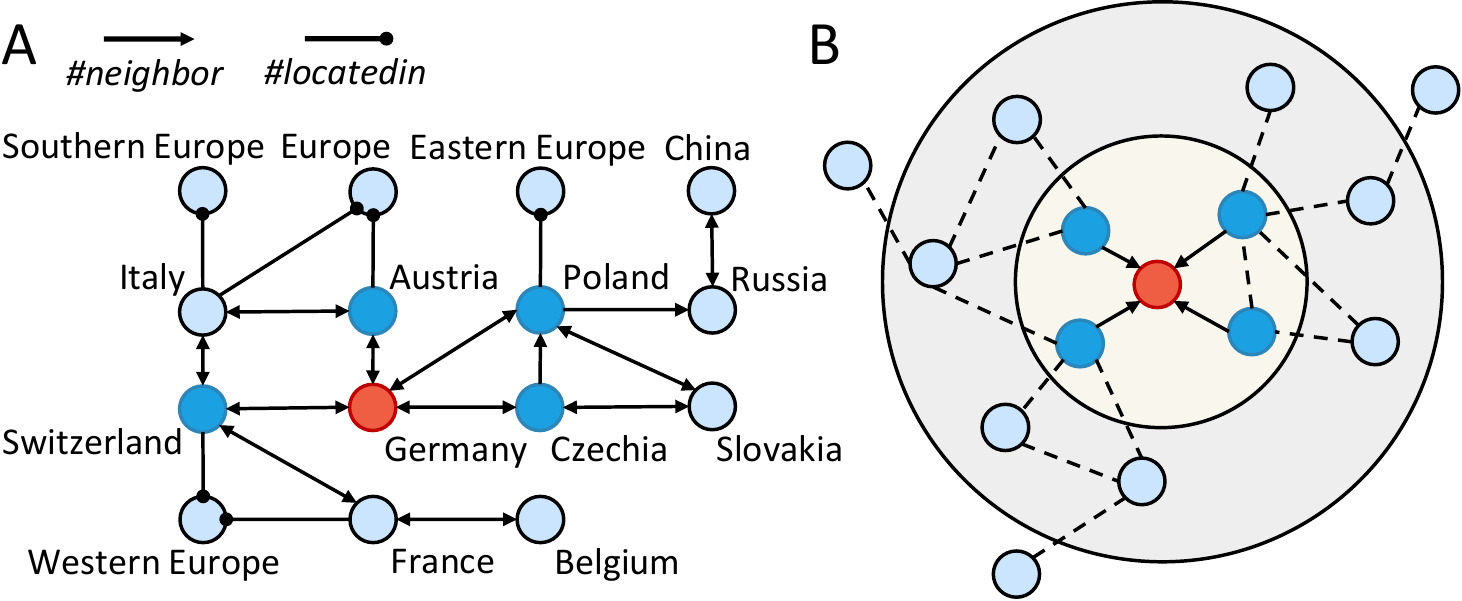}
\caption{\textbf{(A)} Example of a KG with two relations modeling the geographical structure of countries.
Here, nodes represent countries, continents and continental regions, while relations indicate whether two countries are neighbors or whether a country is located in a certain continent or region.
KGs are usually not complete and missing facts have to be inferred from the available graph structure, e.g., whether France and Germany are neighbors, or whether Belgium and Germany are located in Europe.
\textbf{(B)} In GNNs, a node's embedding (red) is enriched by merging it with the embedding of neighboring nodes (blue), hence introducing information about a node's local neighborhood in its vector representation. By repeating this process $L$ times for all nodes, node embeddings accumulate information about other nodes that are at maximum $L$ hops away.}\vspace{-4mm}
\label{fig:KGs}
\end{figure}

KGs structure information in a unified, machine-readable format to represent relational knowledge. Thereby, nodes correspond to entities of the real-world and typed edges between pairs of nodes indicate their relationships and encode factual statements (\cref{fig:KGs}A). While some modern KGs are massive in size, most KGs exhibit incompleteness meaning that not all true facts are contained in the knowledge base. Thus, a popular learning task on KGs is concerned with deriving new facts based on observed connectivity patterns (knowledge base completion, KBC). 

While classical KG reasoning methods employ logical reasoning techniques, scalability issues and breakthroughs of data-driven machine learning methods on other data modalities gave rise to KG reasoning that follow the representation learning paradigm. The basic idea is to embed both entities and relations into low-dimensional vector spaces and model the truthness of facts via functionals on the embedding spaces (see \cite{bordes2013translating, nickel2011three, yang2014embedding, trouillon2016complex}). From an encoder-decoder perspective, earliest KG reasoning models employed shallow embedding lookup as the encoder. More recently, KG reasoning methods that use graph neural networks (GNNs) as an encoder achieved state-of-the-art performance \cite{schlichtkrull2018modeling, vashishth2019composition}. The underlying rationale consists in producing more expressive entity embeddings via pooling information from neighboring entities (\cref{fig:KGs}B).

\begin{figure}[t]
    \centering
    \includegraphics[width=\columnwidth]{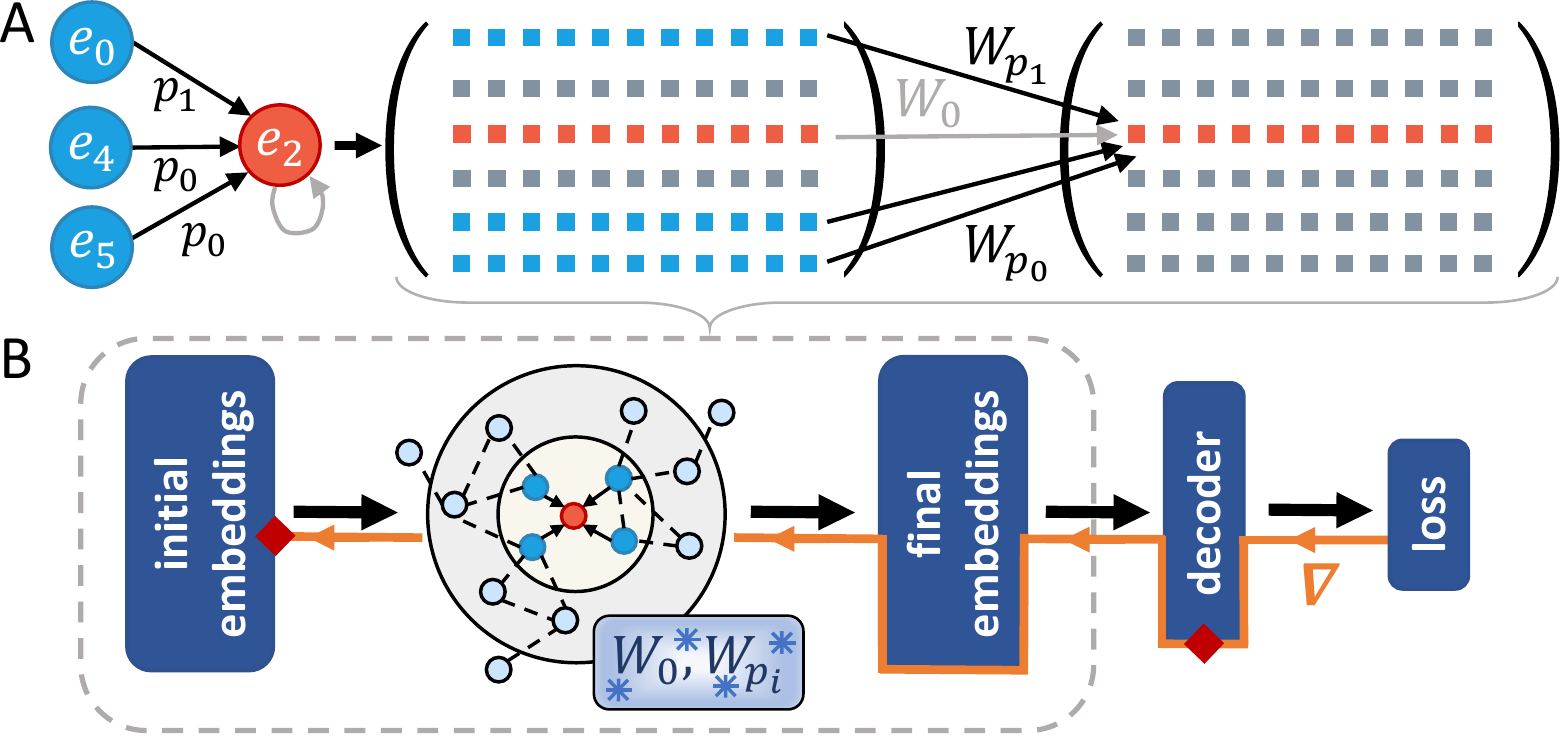}
	\caption{\textbf{(A)} In a GNN, a node's vector representation (red) is updated by accumulating the neighboring node's embeddings (blue), applying a linear transform $W_{p_i}$ and summing over the resulting vectors.
	In R-GCNs, the transform $W_{p_i}$ depend on the relation type $p_i$.
	Optionally, a self-loop can be added (gray, $W_0$), i.e., a node's own embedding is also taken into account when updating it.
	\textbf{(B)} The proposed frozen R-GCN architecture. Weights in the R-GCN layer are kept constant, but both the decoder and -- by propagating an error through the frozen R-GCN (orange) -- the initial embeddings are updated during training (red squares). \vspace{-4mm}
	}	
	\label{fig:frgcn}
\end{figure}

In most state-of-the-art models, for this pooling operation, to acquire a new embedding for a node, the embeddings of its neighboring nodes are aggregated, linearly transformed using a convolutional filter mask and subsequently averaged (\cref{fig:frgcn}A).
Compared to simple lookup encoders, this aggregation-based encoder allows embedding previously unseen nodes \cite{hamilton2017inductive} as well as masking of a node's neighborhood to produce sub-graphs that act as explanations for the output of the decoder \cite{ying2019gnnexplainer,lucic2021cf}.
However, the pooling operator used in GNNs introduces weight sharing during training, i.e., updates to the weights are non-local, which is in stark contrast to the distributed and local design philosophy of neuromorphic hardware.
Additionally, in multi-relational settings, GNNs struggle with overfitting due to the large amount of hyperparameters introduced by the convolutional weights \cite{schlichtkrull2018modeling}.

To solve these challenges, we introduce a frozen R-GCN architecture where the convolutional weights are randomly initialized and kept constant (i.e., frozen) at all times. During training, we only tune the parameters of a task-specific decoder and -- by letting the gradients flow through the static R-GCN -- the initial node embeddings (\cref{fig:frgcn}B).
This way, the model is optimized towards the static filter masks and learns to utilize the aggregation structure to produce richer node embeddings.
Using such a frozen architecture allows us to harness the functional benefits of the aggregation operation while eliminating weight sharing and reducing the amount of trainable parameters in the model.

Bringing such models to neuromorphic hardware, which promise energy efficiency and low times to solution \cite{akopyan2015truenorth,davies2018loihi,wunderlich2019demonstrating,billaudelle2019versatile,rubino2019ultra,mayr2019spinnaker,göltz2020fast}, has the potential of opening a plethora of novel applications and use-cases for these devices -- especially since often recorded data, e.g. in industrial projects \cite{ringsquandl2015semantic,ringsquandl2017event,hubauer2018use,hildebrandt2018configuration,soler2021graph}, has no natural representation as spikes, but can be modelled as heterogeneous graphs.
We are confident that the frozen R-GCN can be mapped to neuromorphic devices as a static structure that allows sensible, local accumulation of information from graph data.
Furthermore, to move closer to the architecture of actual neuromorphic hardware, we combine the frozen R-GCN with a recently proposed spike-based algorithm for shallow graph embeddings \cite{dold2021spikeembed}, extending it to inductive settings like the handling of dynamic graphs.
An intriguing hallmark of this spike-based relational graph convolutional network (SR-GCN) is the simultaneous and highly sparse calculation of initial and final embeddings as time unfolds -- different from non-spiking R-GCNs where embeddings have to be calculated layer by layer.
Moreover, already on conventional computing hardware, for instance CPUs, the suggested frozen architecture shows several benefits like reduced memory footprints and considerable speedups of training without significant performance losses.

The contributions of this work are summarized as follows:
\begin{itemize}
\item We propose and benchmark a frozen R-GCN architecture for efficient calculation of expressive graph embeddings.
\item We experimentally show that the frozen R-GCN structure reduces memory and compute time without significant performance losses.
\item We show that the  deep graph encoder  can  be  combined  with  arbitrary  shallow decoder models,  which  we  demonstrate  by  constructing  the  first spike-based R-GCN.
\item  The spiking model constitutes a novel way of representing and calculating graph embeddings in a purely temporal and event-based way.
\end{itemize}
In the remaining sections, we first introduce the mathematical notation used throughout the paper as well as the background required to follow this work in \cref{sec:backround_and_notation}. Subsequently, in \cref{sec:frozen_graph_convolutions}, we propose the frozen R-GCN and evaluate the KBC performance on the benchmark data sets FB15k-237, UMLS, and Countries S1. \cref{sec:spike-based_graph_embedding} outlines first proof of concepts of spike-based R-GCNs for KG reasoning before we summarize our results and conclude in \cref{sec:summary,sec:conclusion}.

%% file: content/background.tex
Before proceeding, we first define the mathematical notation that we use \sloppy{throughout} this work and provide the necessary background on KGs.

\subsection{Notation}
Scalars are indicated by lower case letters ($x \in \mathbb{R}$), column vectors by bold lower case letters ($\pmb{x} \in \mathbb{R}^n$), and matrices by upper case letters ($X \in \mathbb{R}^{n_1 \times n_2}$). Moreover, three-way tensors are denoted by calligraphic letters  ($\mathcal X \in \mathbb{R}^{n_1 \times n_2\times n_3}$).
Sets are either indicated by their canonical symbols (e.g., $\mathbb{N}$ denotes the set of natural numbers) or by calligraphic letters.
$\dot{x}$ is the time derivative $\frac{\text{d}x}{\text{d}t}$ of $x$.

\subsection{Knowledge graphs}

KGs are collections of factual statements that specify the relations between entities of the real world. We denote with $\mathcal{E}$ the set of relevant entities. $\mathcal{R}$ denotes the set of binary relations. In this work, a KG is defined as a collection of triples $\mathcal{KG} \subset \mathcal{E} \times \mathcal{R} \times \mathcal{E} $. Each triple in $\mathcal{KG}$ corresponds to a factual statement $(s,p,o)$ –  where $s$ indicates the subject, $p$ the predicate, and $o$ the object. While every triple in $\mathcal{KG}$ is interpreted as a true fact, there exist different interpretations of absent triples. Since large scale KGs are typically incomplete, it is common to make the open world assumption (OWA). The OWA states that one cannot conclude that absent triples are false -- their truth value is rather unknown. In this setting, KBC is a typical learning task related to KGs.

Data-driven KBC techniques are studied under the umbrella term statistical relational learning (SRL) \cite{nickel2015review}. Among these methods KG embeddings have become the dominant approach. Thereby, both entities and relations are projected into low-dimensional vector spaces encoding connectivity patterns between entities. In these embedding spaces, the interactions between the embeddings of entities and relations can be efficiently modelled via functionals to produce scores that indicate the likelihood of triples. KG embedding methods can be categorized according to fundamental interaction mechanisms of the functionals. For example, translational models such as TransE \cite{bordes2013translating} embed both entities and relations into the same vector space and model the action of different relations as vector space translations. Concretely, TransE imposes
\begin{equation}
    \ent{s} + \vrel{p} \approx \ent{o} \ \ \ \ \text{if} \ (s,p,o) \in \mathcal{KG}\,,
\end{equation}
where the bold letters correspond to $d-$dimensional vector space embeddings of the corresponding entities and relations. During training, these embeddings are tuned such that the discrepancy between $\ent{s} + \vrel{p}$ and $\ent{o}$ (measured by some metric on $\mathbb{R}^d$) serve as a proxy for the plausibility of triples. 

Multiplicative models implicitly correspond to tensor decomposition models where the score of each fact is given by a bilinear form. Note that KGs have a natural representation in terms of adjacency tensors $\mathcal{X} \in \{0,1 \}^{n_E \times n_E \times n_R}$. Similar to an adjacency matrix of a homogeneous graph, an entry of $\mathcal{X}$ indicates the absence (0) or presence (1) of a triple. Various multiplicative KG reasoning models correspond to different formulations of the bilinear forms. The tensor factorization model RESCAL \cite{nickel2011three} considers bilinear forms induced via relation-specific, quadratic matrices. However, this leads to one of the main disadvantages of RESCAL: the number of parameters grows quadratically in the embedding space dimension. As a remedy to this problem, the quadratic matrices of RESCAL are constrained to be diagonal, which is generally known as DistMult \cite{yang2014embedding}. Concretely, DistMult scores triples via 
\begin{equation}
    \label{eq:distmult}
    \mathcal{X}_{s,p,o} \approx  \entT{s} \text{diag}(\vrel{p}) \ent{o} \,,
\end{equation}
where $\ent{s}, \vrel{p}, \ent{o} \in \mathbb{R}^d$ and $\text{diag}(\vrel{p})$ is a diagonal matrix with diagonal entries given by $\vrel{p}$.

\subsection{Graph neural networks}

GNNs are neural networks that aim to produce expressive representations of the nodes in homogeneous graphs via a message passing heuristic between neighboring nodes (see  \cite{scarselli2008graph}). One of the most influential methods that led to a widespread popularization of GNNs is the graph convolutional network (GCN) introduced by  \cite{kipf2016semi}. Similar to convolutional neural networks (CNNs) on regular grids (e.g., images or time series), GCNs aim to extract localized features by aggregating information from different neighborhoods in the graph via the same filtering operations. This imposes not only location invariant feature mappings but also leads to parameter sharing and an efficient regularization effect. Recently, GNNs have not only been applied to classical graph learning tasks but have also achieved state-of-the-art performance on various data modalities (e.g., \cite{shen2018person,yao2019graph}). Moreover, and most relevant for this work, there have been attempts to generalize GNNs to KGs. The GNN model that is most relevant to this work is the R-GCN \cite{schlichtkrull2018modeling}. The underlying idea of R-GCNs is to process the embeddings of neighboring entities via relation-specific linear mappings, pool this information, and combine it with the center node embedding to update the center node embedding. Concretely, for a center node $i \in \mathcal{E}$, we have that the embedding after layer $l$ of the R-GCN is given by
\begin{equation}
\label{eq:rgcn_encoder}
 \ent{i}^{(l+1)}= \phi \left( \sum_{p \in \mathcal{R}}  \sum_{j \in \mathcal{N}_i^p} \frac{1}{|\mathcal{N}_i^p|} W_p^{(l)} \ent{j}^{(l)} + W_0^{(l)} \ent{i}^{(l)} \right ) \, ,
\end{equation}
where $\phi$ is a non-linearity, $\mathcal{N}_i^p$ denotes the graph neighborhood of node $i$ with respect to relation $p$ (i.e., $\mathcal{N}_i^p = \left\{j \in \mathcal{E} \vert (i,p,j) \in \mathcal{KG} \right\}$) and $|\mathcal{N}_i^p|$ is the number of elements in $\mathcal{N}_i^p$.
Moreover, $\ent{i}^{(l)} \in \mathbb{R}^d$ with $l\geq 1$ denotes the embedding of entity $i$ produced by the $l$-th layer of the R-GCN and $\ent{i}^{(0)}$ the initial embedding. $W_p^{(l)}, W_0^{(l)} \in \mathbb{R}^{d \times d}$ correspond to trainable weight matrices that act on the embeddings of neighboring nodes and the center node, respectively.  Note that while the first summand $\sum_{p \in \mathcal{R}}  \sum_{j \in \mathcal{N}_i^p} \frac{1}{|\mathcal{N}_i^p|} W_p^{(l)} \ent{j}^{(l)}$ pools the information from neighboring nodes, the second summand $W_0^{(l)} \ent{i}^{(l)}$ is a transform of the center node's own representation in the previous layer (\cref{fig:frgcn}A). Thus, the second term corresponds to a self-loop and, depending on the intended usage, can be omitted, e.g., to enable the R-GCN to operate in an inductive setting. Multiple layers corresponding to \cref{eq:rgcn_encoder} can be stacked on top of each other to increase the receptive field. In particular, in a R-GCN with $L$ layers each center node receives information from the entities $L$ hops away (\cref{fig:KGs}B). In order to address various tasks, R-GCNs can be composed with task-specific decoders. For example, for the KBC task,  \cite{schlichtkrull2018modeling} compose a R-GCN encoder with the scoring function of DistMult, i.e., triples are scored according to \cref{eq:distmult}.

%% file: content/FRGCN.tex
\subsection{Method}
Based on the observation of \cite{kipf_2016} that GCNs with randomly sampled weights are able to produce meaningful node embeddings that preserve local neighborhood structures of homogeneous graphs, we propose to compose an untrained R-GCN with an arbitrary link prediction model. Concretely, we call a R-GCN model frozen if the weight matrices, i.e., $W_p^{(l)}$ and  $W_0^{(l)}$ in \cref{eq:rgcn_encoder}, are initialized according to some (possibly random) law but are subsequently not tuned any more to fit the training data. The frozen R-GCN constitutes an entity encoder that takes as input the initial entity representations and computes a neighborhood-aware embedding. To ease the notation, we denote with $\pmb{f}^{\text{frzn}}: \mathcal{E} \rightarrow \mathbb{R}^d$ the mapping induced by $L \in \mathbbm N$ layers of a frozen R-GCN computed according to \cref{eq:rgcn_encoder}. Note that we consider formulations of the R-GCN layer both with ($W_0 \neq 0$) and without ($W_0 = 0$) self-loops. Thereby, the initial features of an entity are given by 
\begin{equation}
    \ent{s}^{(0)} = E  \mathbbm{1}_s,
\end{equation}
where $E \in \mathbb{R}^{d \times \vert \mathcal E \vert}$ is a matrix that contains trainable entity embeddings and $\mathbbm{1}_s \in \mathbb{R}^{\vert \mathcal E \vert}$ is a vector of zeros except for a one at the position corresponding to the index of entity $s$. Subsequently, in order to compute scores for the plausibility of a triple, we feed the entity representation resulting from $\pmb{f}^{\text{frzn}}$ into the scoring function of TransE. This leads to
\if false
In this work we consider the three models DistMult, ComplEx, and TransE reviewed in \cref{sec:backround_and_notation}. Concretely, composing DistMult with the frozen R-GCN to score a triple $(s,p,o) \in \mathcal{E} \times \mathcal{R} \times \mathcal{E}$, leads to
\begin{equation}
   S(s,p,o) =  f_{\text{frzn}}(s)^\intercal \text{diag}(\mathbf{p}) f_{\text{frzn}}(o) \,,
\end{equation}
where $\mathbf{p}$ is a trainable representation of $p$ produced via a simple embedding lookup
\begin{equation}
\label{eq:lookup_p}
    \mathbf{p} = R  \mathbbm{1}_p \, 
\end{equation}
where $R \in \mathbb{R}^{d \times \vert \mathcal R \vert}$ is an embedding matrix for all relations. Similarly, using ComplEx for scoring triples, leads to
\begin{equation}
    S(s,p,o) = 
   \text{real}\left( f_{\text{frzn}}(s)^\intercal \text{diag}(\mathbf{\bar p}) \mathbf{\overline{f_{\text{frzn}}(o)}}\right) \, .
\end{equation}
Then, up to additive regularization terms, the training objective is given by a least squares criterion
\begin{equation}
    \min_{\theta} \sum_{(s,p,o)\in \mathcal{T}} \left( \mathcal{X}_{s,p,o}- S(s,p,o) \right)^2 \, ,
\end{equation}
where $\mathcal{T}$ is a training set that contains next to observed triples (i.e., $(s,p,o) \in \mathcal{KG}$ with $\mathcal{X}_{s,p,o} = 1$) also negative triples ($\mathcal{X}_{s,p,o} = 0$). $\theta = \{ E, P \}$ denotes the set of trainable parameters.
\fi
\begin{equation}
    d(s,p,o) = \norm{\pmb{f}^{\text{frzn}}(s) + \vrel{p} - \pmb{f}^{\text{frzn}}(o)} \, ,
\end{equation}
where $\norm{\cdot}$ denotes the L1 norm. We also experimented with other scoring functions like RESCAL, DistMult, or ComplEx \cite{trouillon2016complex}, but found that TransE yields the best performance.

$\vrel{p}$ is a trainable representation of relation type $p$ produced via an embedding lookup
\begin{equation}
\label{eq:lookup_p}
    \vrel{p} = R  \mathbbm{1}_p \,,
\end{equation}
where $R \in \mathbb{R}^{d \times \vert \mathcal R \vert}$ is an embedding matrix for all relations. Then, up to additive regularization terms, the training objective is given by the hinge loss
\begin{equation}
\label{eq:frozen_loss}
     \min_{\theta} \sum_{\substack{(s,p,o) \in \mathcal{T}_+ \\ (\tilde s,p, \tilde o) \in \mathcal{T}_-}} \left[\gamma +  d(s,p,o) - d(\tilde{s},p,\tilde{o}) \right]_+ \,,
\end{equation}
where $[x] = \max(x,0)$ and $\gamma \geq 0$ is a hyperparameter that determines the margin of the hinge loss. Furthermore, $\mathcal{T}_+ \subset \mathcal{KG}$ is the set of observed training triples and $\mathcal{T}_- \subset \mathcal{E} \times \mathcal{R} \times \mathcal{E}$ is a set of negative triples where each element is obtained by substituting either the subject or the object entity from an observed triple in $\mathcal{T}_+$. $\theta = \{ E, R \}$ denotes the set of trainable parameters and it is crucial for our method that the weight matrices of the R-GCN are not contained here. However, during the backward pass, the gradient of the supervision signal produced by \cref{eq:frozen_loss} is propagated through  $\pmb{f}^{\text{frzn}}$ to tune the entity embeddings $E$ (see \cref{fig:frgcn}), allowing them to learn how to utilize the frozen R-GCN structure.

\begin{table}[b]
\caption{Statistics of the data sets used in the experiments with the frozen R-GCN.}
\label{tab:datasets}
\center
\begin{tabular}{ c|  c|  c | c}
 Data set & Entities & Relations & Triples \\
 \hline \hline
FB15k-237 & 14,541 & 237 & 310,116 \\
UMLS & 135 & 49 & 5,216\\
Countries S1 & 272 & 2 & 1,158\\
\hline\hline
\end{tabular}
\end{table}

\input{content/table1}
\begin{figure*}[ht]
    \centering
    \includegraphics[width=2\columnwidth]{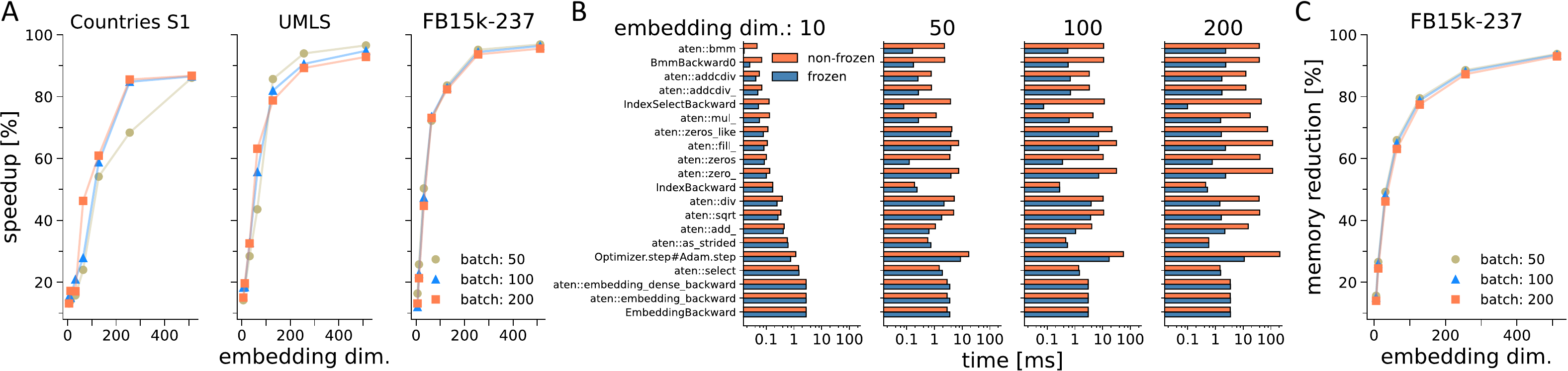}
	\caption{Comparison of speed and memory consumption of the backward pass and optimizer step for the frozen (frzn) and non-frozen (non-frzn) architecture.
	\textbf{(A)} Speedup $(\Delta t^\text{non-frzn} - \Delta t^\text{frzn}) / \Delta t^\text{non-frzn}$ of calculating and applying gradients.
	Each data point is an average of 100 separate backward passes.
	Error bars are negligible and hence not shown to increase readability.
	\textbf{(B)} Most time intensive function calls during a training step. We show the full time required by an operation here, i.e., including the time required for further function calls in an operation. Recorded with PyTorch's profiler.
	\textbf{(C)} Memory reduction for training with frozen architecture. 
	We record the memory consumption $m$ of the most memory intensive operation in the backward pass and show $(m^\text{non-frzn} - m^\text{frzn}) / m^\text{non-frzn}$. 
    Recorded with PyTorch's profiler.
	Measurements were done using an off-the-shelf Intel Core i9-9900KF@3.6GHz processor.
	}\vspace{-1mm}	
	\label{fig:profiler}
\end{figure*}

\subsection{Experiments}
In what follows we detail an empirical study to (i) evaluate the performance of the proposed method on the KBC task and (ii) compare the running time and memory consumption of the frozen R-GCN architecture with a model that optimizes all weight parameters. 

\paragraph{Data sets}
We evaluate our method on three different data sets: Countries S1, UMLS, and FB15k-237. \cref{tab:frozen_rgcn} contains the most important summary statistics on the sizes of the three data sets.  Countries is a carefully designed data set to examine the reasoning abilities of KBC models. Thereby, the entities correspond to either countries, regions, or subregions and the task is to infer geographic relations. UMLS is a biomedical KG holding facts about diseases, chemical compounds, and their relations. FB15k-237 is a general-purpose KG extracted from the bigger data set FB15k to prevent leakage between the training and the test set, making the data sets more challenging. 

\paragraph{Experimental protocol and metrics}
We compare the performance of the frozen R-GCN on the KBC task with the shallow KG embedding methods TransE and DistMult as well as the original R-GCN model that uses DistMult as a decoder. Moreover, we also evaluate the performance of a R-GCN model composed with TransE, where all parameters -- including the weight matrices in \cref{eq:rgcn_encoder} -- are trained in an end-to-end fashion (with and without self-loops). The hyperparameters of all considered methods are specified in the supplementary material of this work.
 
We adopt the standard ranking-based procedure proposed by \cite{bordes2011learning}. Concretely, for each test triple $(s,p,o)$ we remove either the subject (object) entity to create a query $(s,p,?)$ ($(?,p,o)$). Subsequently, all entities $e \in \mathcal{E}$ that do not correspond to observed subject/object entities (i.e., $(s,p,e) \in \mathcal{KG}$ and $(e,p,o) \in \mathcal{KG}$) substitute the placeholder in the query and the resulting candidate triples are scored via a KBC model. These scores are used to rank all entities and the different models are evaluated by their ability to rank the original triples as high as possible, i.e., in the best case we rank the original triple $(s,p,o)$ at the first position. To compare the performance of different methods, we use the standard performance measures mean reciprocal rank (MRR; the average of the inverse ranks), the hits@1, and hits@3, i.e., the proportion of test triples that is ranked as the top triple or among the top three triples, respectively.

\paragraph{Results} Table \ref{tab:frozen_rgcn} summarizes the findings of the experimental study with the frozen R-GCN. On the largest data set FB15k-237, we have that the frozen R-GCN with self-loops achieves the best performance among all considered methods with respect to all metrics. In particular, the frozen R-GCN with self-loops outperforms both the shallow TransE model and a R-GCN with tuned weight matrices. Based on the rather low  performance of the frozen R-GCN without self-loops, we can see that including information of the center node from the previous layer is an essential feature that leads to a significant performance boost. The results on the biomedical KG UMLS follow the same pattern previously described on FB15-237. Again, the frozen R-GCN shows the best performance with respect to all metrics (the shallow TransE model achieves the same value for hits@3). On the smallest data set that we considered in this study, Countries S1, the results are more ambiguous in the sense that the fully trained R-GCN with self-loops outperforms the frozen R-GNC with respect to MRR and hits@1. Moreover, the shallow TransE model outperforms all other considered methods with respect to hits@3 by achieving a perfect score of 1. Overall, across all data sets, we find that the frozen R-GCN model with self-loops can keep up with the performance of all considered baseline methods. In particular, on the larger data sets UMLS and FB15k-237, the frozen R-GCN outperforms both the fully trained R-GCN and the shallow TransE model.

We further measured the running time during training and the memory consumption of both the frozen R-GCN and the fully trained model. The most relevant results are shown in Figure \ref{fig:profiler}. For example, we find that when averaging over the running times of 100 backward passes, the frozen architecture leads to speedups of more than 90\% compared to the fully tuned model. Thereby, the relative speedup is more pronounced the larger the training set and the embedding size. Similarly, the memory reduction of the frozen model reaches more than 90\% on FB15k-237 as the embedding size increases.
The speedup and memory reduction are mostly due to less demanding operations in the backward pass, since the gradients with respect to the GNN weights are not calculated and stored when freezing the weights.
In \cref{fig:profiler}B, we list the most time-consuming operations of the backward pass.
For small embedding dimensions, the differences are negligible, but for higher dimensions, operations like batch matrix multiplication (bmm) take considerably longer if weights are not frozen.

\begin{figure}[t]
    \centering
    \includegraphics[width=\columnwidth]{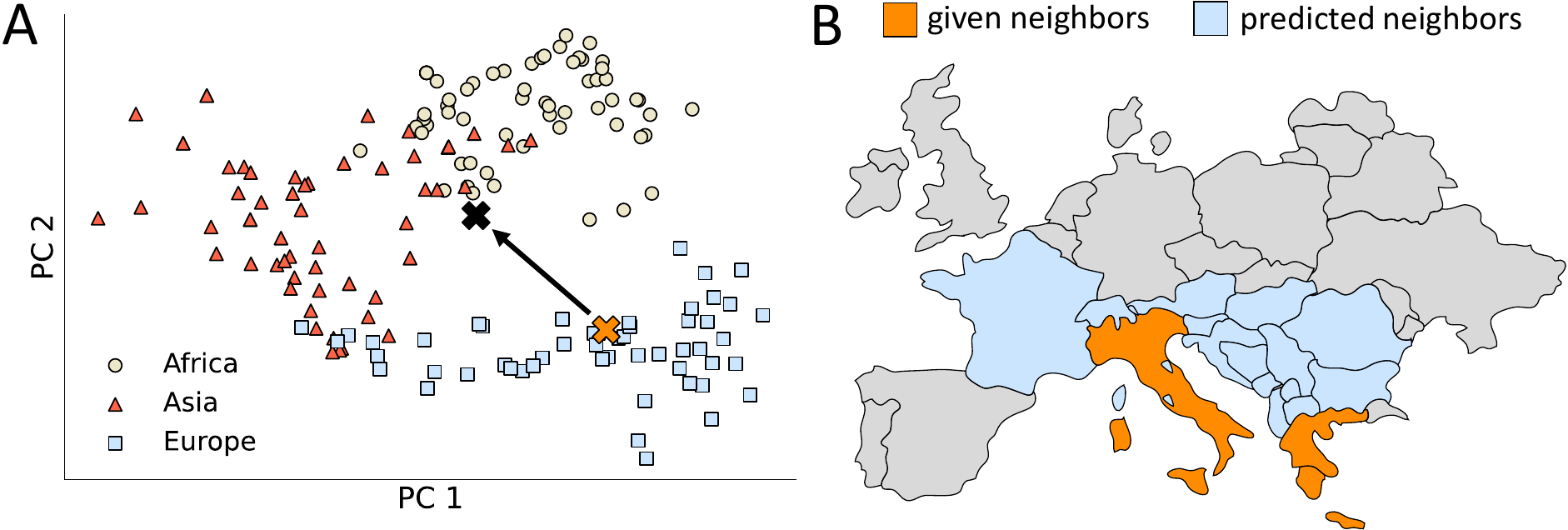}
	\caption{Illustration of the inductive KBC setting on the Countries data set. \textbf{(A)} Visualisation of node embeddings using the first two principal components.
	Novel nodes can be embedded by knowing part of their neighborhood alone (orange and black cross).
	\textbf{(B)} Given the neighbors of a novel entity (orange), the frozen R-GCN predicts sensible suggestions for potential neighbors (blue).}	
	\label{fig:inductive}
\end{figure}

Next to experiments in the canonical, transductive KBC setting, we also perform a qualitative evaluation of the inductive reasoning capabilities of the proposed frozen architecture. Concretely, we want to analyze  whether our model can compute meaningful predictions when the subject entity is not encountered during training. This setting is challenging, because the embedding of the subject is not tuned during training and all relevant information needs to be obtained via pooling from the neighboring nodes. Note that without the self-loop in \cref{eq:rgcn_encoder}, the frozen R-GCN does not rely on the embedding of the center node but only on the embeddings of its neighbors. In this experiment, we assume that all entities in the receptive field are known and only the center node is novel. Due to its intuitive nature, we consider a modified version of the Countries data set. Concretely, we add a new entity (i.e., a new country that does not exist) to the graph with the only provided neighbors being Italy and Greece. \cref{fig:inductive}A shows a visualisation of the node embeddings using the first two principal components. It is apparent that the new entity is placed close to other countries situated in Europe (orange cross). 
A closer look at the neighborhood of the novel node can be obtained by ranking all other countries using our model and, depending on the scores, creating a list of plausible neighbors.
In \cref{fig:inductive}B, we show the highest ranking countries in blue. As expected, most predicted neighbors are located between Italy and Greece.
Furthermore, by changing the new country's neighborhood, e.g., by replacing Italy with Egypt, the embedding starts moving in the embedding space, placing the new node into a region that is more consistent with the adjusted neighborhood (\cref{fig:inductive}A, black cross).

To summarize, we observe that even untrained, i.e., randomly initialized and frozen, GNN architectures are sufficiently structured (aggregate, filter and average) to be used as sensible feature selectors.
Consequently, during training, the initial entity embeddings align themselves accordingly with these static feature masks to achieve high performance on KBC tasks -- as seen in the previously presented experiments.

%% file: content/table1.tex
\begin{table*}[htbp]
\caption{Test performances of frozen R-GCNs with baseline comparisons. Best values shown in bold.}
\label{tab:frozen_rgcn}
\begin{center}\renewcommand{\arraystretch}{1.2}
\begin{tabular}{c | c c c | c c c | c c c }
\multicolumn{1}{c}{}&\multicolumn{3}{c}{FB15k-237}&\multicolumn{3}{c}{UMLS}&\multicolumn{3}{c}{Countries S1} \\
\hline\hline
Model & MRR& hits@1& hits@3 & MRR& hits@1& hits@3 & MRR& hits@1& hits@3 \\
\hline\hline
TransE \cite{bordes2013translating}
 &  0.24 & 0.15 & 0.27 & 0.78 & 0.60 & \best{0.94} & 0.87 & 0.75 & \best{1.00}     \\
 & & & & & & & & & \\
\hline
DistMult \cite{yang2014embedding}
 &  0.15 & 0.08 & 0.16 & 0.64 & 0.53 & 0.69  & 0.49 & 0.35 & 0.50 \\
 & & &  & & & & & &\\
\hline
R-GCN + DistMult \cite{schlichtkrull2018modeling} 
 &  0.20 & 0.09 & 0.23 & 0.44 & 0.33 & 0.48 & 0.38 & 0.23 & 0.50  \\
 & & & & & & & & &\\
\hline
\hline
R-GCN + TransE 
 &  0.23 & 0.14 & 0.25 & 0.58 & 0.37 & 0.74  & 0.79 & 0.67 & 0.94 \\
 w/ self-loop & & & & & & & & & \\
\hline
R-GCN + TransE  
&  \best{0.26} & \best{0.17} & \best{0.29} & \best{0.80} & \best{0.65} & \best{0.94} & 0.90 & 0.83 & 0.98   \\
frozen + w/ self-loop & & & & & & & & &\\
\hline\hline
R-GCN + TransE  
&  0.21 & 0.13 & 0.22 & 0.76 & 0.64 & 0.85 & \best{0.95} & \best{0.92} & 0.98   \\
w/o self-loop& & & & & & & & & \\
\hline
R-GCN + TransE 
& 0.14 & 0.08 & 0.14 &  0.62 & 0.48 & 0.70 & 0.77 & 0.65 & 0.88   \\
frozen + w/o self-loop & & & & & & & & & \\
\hline \hline
\end{tabular}
\label{tab1}
\end{center}
\end{table*}

%% file: content/SRGCN.tex
The modular structure of R-GCNs enables us to combine the proposed frozen structure with arbitrary shallow embedding methods.
Hence, to move the proposed architecture closer to current iterations of neuromorphic hardware -- which mostly implement spiking neurons -- we investigate a spike-based version of R-GCNs based on the shallow graph embedding model SpikE introduced in \cite{dold2021spikeembed}.

\subsection{Shallow embedding model}
\input{content/table2}

\begin{figure}[t]
    \centering
    \includegraphics[width=0.85\columnwidth]{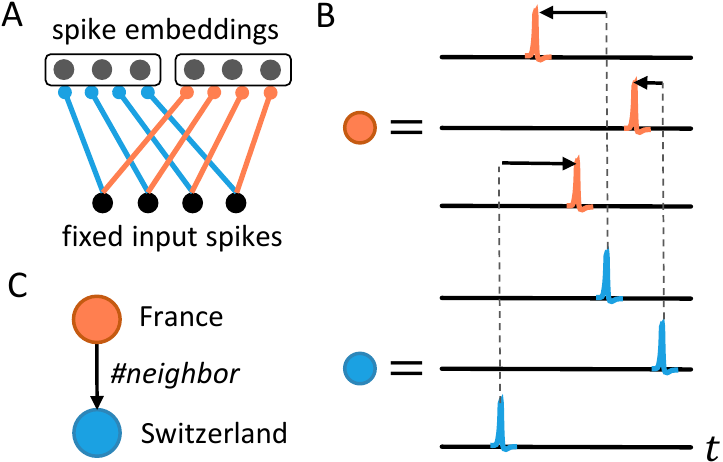}
	\caption{\textbf{(A)} Network architecture for spike-based graph embedding, with a fixed input layer and an embedding layer.
	During training, weights (blue and red) are adjusted to change the spike embeddings.
	\textbf{(B)} Nodes in a graph are represented as first-spike times of neuron populations, while relations are encoded as expected spike time differences between populations. 
	\textbf{(C)} Example graph corresponding to (B), i.e., where node and relation embeddings match.
	\vspace{-4mm}
	}	
	\label{fig:nLIF}
\end{figure}

A natural way of mapping the symbolic structure of graphs to spiking neural networks is by representing nodes as spike times of neuron populations and relations as spike time differences between populations.
Similar to \cite{dold2021spikeembed}, we represent a node $s$ in the graph by the first spike times $\stime{s} \in \mathbb{R}^N$ of a population of $N \in \mathbb{N}$ integrate-and-fire neurons (nLIF) with exponential synaptic kernel $\kappa(x,y) = \heaviside{x-y} \exp\left(-\frac{x-y}{\taus}\right)$,
\begin{equation}\label{eq:dotu}
    \dot{u}_{s,i}(t) = \frac{1}{\taus}\sum_{j} w_{s,ij} \, \kappa(t, t^\mathrm{I}_j) \,,
\end{equation}
where $u_{s,i}$ is the membrane potential of the $i$th neuron of population $s$, $\taus$ the synaptic time constant and $\heaviside{\cdot}$ the Heaviside function.
A spike is emitted when the membrane potential crosses a threshold value $\thresh$.
$w_{s,ij}$ are synaptic weights from a pre-synaptic neuron population, with every neuron $j$ emitting a single spike at fixed time $t^\mathrm{I}_j$ (\cref{fig:nLIF}A).

Similarly, relations are encoded by a $N$-dimensional vector of spike time differences $\pmb{\Delta}_p \in \mathbb{R}^N$.
Whether a triple \triple{s}{p}{o} is plausible or not is evaluated by looking at the discrepancy between the spike time differences of the node embeddings, $\stime{s} - \stime{o}$, and the relation's embedding $\pmb{\Delta}_p$ (\cref{fig:nLIF}B):
\begin{equation}\label{eq:spikedecoder}
    d(s,p,o) = \lone{\stime{s} - \stime{o} -\pmb{\Delta}_p} \,,
\end{equation}
where $\norm{\cdot}$ is the L1 norm.
If a triple is valid, then the patterns of node embeddings and relation match, leading to $d(s,p,o) \approx 0$, i.e., $\stime{s} \approx \stime{o} + \pmb{\Delta}_p$ (\cref{fig:nLIF}B,C).
If the triple is not valid, we have $d(s,p,o) > 0$, with higher discrepancies representing less plausibility.

Given a KG $\mathcal{KG}\subset \mathcal{E} \times \mathcal{R} \times \mathcal{E}$, suitable spike embeddings are found by minimizing the margin-based ranking loss \cref{eq:frozen_loss}, with $\theta$ now being the weights $w_{s,ij}$, $\forall s \in \mathcal{E}$, of the nLIF embeddings and the relation embeddings $\pmb{\Delta}_p$, $\forall p \in \mathcal{R}$.

\subsection{Hybrid graph convolutional model}

As a first step, we use the spike-based model to obtain initial embeddings $\stime{i}$, which are subsequently fed into the frozen R-GCN architecture with one layer to produce more expressive spike embeddings
\begin{equation}
\label{eq:frgcn_hybrid}
 \pmb{f}^\mathrm{frzn}(i)= \sum_{p \in \mathcal{R}}  \sum_{j \in \mathcal{N}_i^p} \frac{1}{|\mathcal{N}_i^p|} W_p \stime{j} + W_0 \stime{i} \,.
\end{equation}
The updated spike embedding $\pmb{f}^\mathrm{frzn}(i)$ of node $i$ is given by the weighted average of its neighboring nodes' and its own spike times.
Since this can be interpreted as feeding spike times into a layer of artificial neurons, we call this version the ``hybrid'' model in this document.
During training, the initial spike times adapt to the fixed weights used for averaging to produce neighborhood-aware embeddings.
This is demonstrated in \cref{tab2}, where the hybrid model achieves similar or better results than the vanilla spike-based embedding model on the UMLS and Countries S1 data sets.
 
A downside of the hybrid model is the locked calculation of the spike time averages, which contradicts the event-based and asynchronous computing paradigm of spiking neural networks.
Hence, in the following, we introduce a fully spiking R-GCN which we call SR-GCN.

\subsection{Spiking graph convolutions}

The default R-GCN structure \cref{eq:rgcn_encoder}, \cref{eq:frgcn_hybrid} is basically a layer of artificial neurons with special routing of the input.
Thus, a fully spiking model can be obtained by replacing the artificial neurons in the R-GCN layer by spiking neurons.
To guarantee consistency with the first layer, i.e., the initial embeddings \cref{eq:dotu}, we again use nLIF neurons, resulting in the following interaction for the SR-GCN
\begin{equation}
\label{eq:sgnn}
 \dot{\pmb{u}}_s^\mathrm{frzn}(t) = \frac{1}{\taus} \sum_{p \in \mathcal{R}}  \sum_{j \in \mathcal{N}_s^p} \frac{1}{|\mathcal{N}_s^p|} W_p \pmb{\kappa}(t,\stime{j}) + W_0 \pmb{\kappa}(t,\stime{s}) \,,
\end{equation}
where $\pmb{\kappa}$ is applied component-wise, i.e., $\pmb{\kappa}(x,\pmb{y})_i = \kappa(x, y_i)$. 
Updated spike embeddings are then obtained by applying the spike condition, e.g., for the $i$'th neuron of population $s$, the time to first spike is calculated via $u_{s,i}^\mathrm{frzn}(t^\mathrm{frzn}_{s,i}) \overset{!}{=} \thresh$.
\input{content/table3}

\begin{figure*}[!htbp]
    \centering
    \includegraphics[width=2\columnwidth]{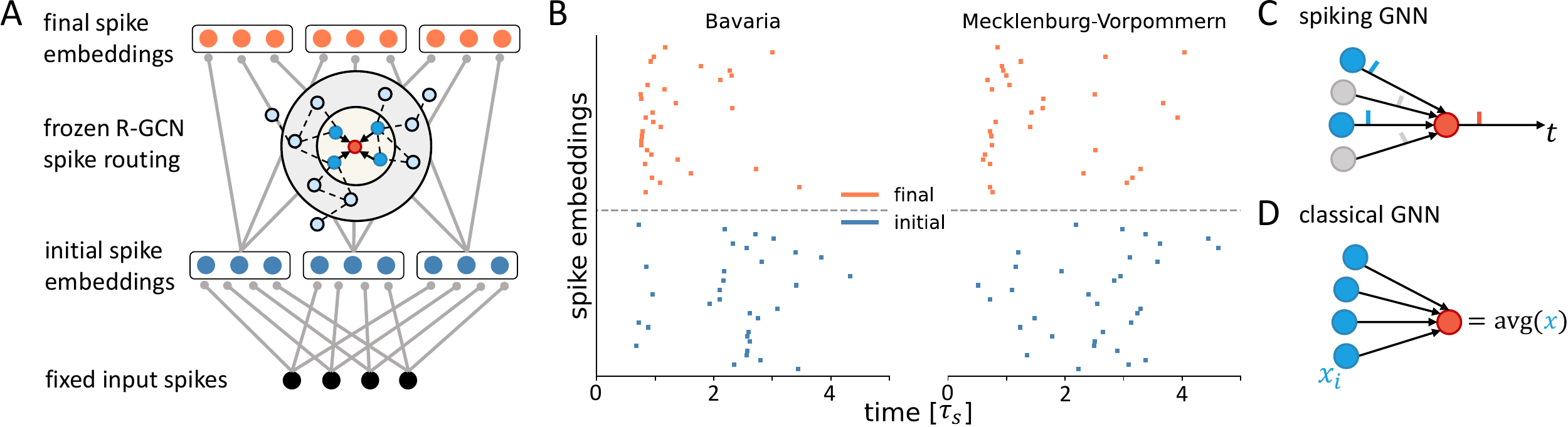}
	\caption{\textbf{(A)} Structure of the SR-GCN.
	\textbf{(B)} Initial (bottom, blue) and final (top, red) embeddings for the two entities Bavaria and Mecklenburg-Vorpommern in the Federal States data set after training. Both initial and final embeddings are calculated simultaneously in an event-based way.
	\textbf{(C)} Message passing in the SR-GCN model is purely event-based and only causal pre-synaptic spikes (causal: blue, non-causal: gray) are considered for calculating the spike times of the final embedding layer (red).
	\textbf{(D)} In contrast, in classical GNN architectures, all components of the initial embeddings are used to calculate final embeddings.
	}	\vspace{-2mm}
	\label{fig:srgcn}
\end{figure*}

The fully spiking model consists of three nLIF layers (\cref{fig:srgcn}A):
\begin{itemize}
    \item An input layer which provides a pool of fixed spike times $\pmb{t}^\mathrm{I}$ (\cref{fig:srgcn}A, bottom).
    \item An initial embedding layer \cref{eq:dotu} with a population for each node in the graph.
    The populations get $\pmb{t}^\mathrm{I}$ as input through trainable weights.
    The initial embedding of node $s$ is given by the vector of spike times $\stime{s}$ of population $s$.
    \item A final embedding layer \cref{eq:sgnn} with a population for each node in the graph (\cref{fig:srgcn}A, top).
    Each population obtains inputs from the initial embedding layer through the frozen R-GCN structure.
    The embedding of node $s$ is given by the vector of spike times $\pmb{t}^\mathrm{frzn}_s$ of population $s$.
\end{itemize}
To train the model, we use the same decoder and loss function as for SpikE \cref{eq:spikedecoder}, \cref{eq:frozen_loss} and optimize both the weights for the initial embeddings and the relation embeddings while keeping the R-GCN weights frozen.

In contrast to classical GNN models, the spike-based version computes embeddings in all layers simultaneously (\cref{fig:srgcn}B).
Furthermore, when accumulating the embeddings of neighboring nodes, only causal input spikes that precede the output spike are taken into account to update the embeddings (\cref{fig:srgcn}C), different from classical GNNs that average and filter the whole embedding vector (\cref{fig:srgcn}D).
This constitutes a novel way of processing graphs, where important information is encoded in early spike times and message passing between nodes in the graph is done in a purely event-based way.

Simulating SR-GCNs is computationally quite demanding, and hence we restrict ourselves here to a proof of concept on smaller data sets that greatly reduce simulation times.
We designed two smaller data sets for our experiments: (i) one based on the famous video game StarCraft: Brood War, consisting of 32 entities, 5 relation types, 65 training triples and 11 evaluation and test triples, and (ii) one modelling the geographic relationships between the federal states in Germany, consisting of 27 entities, 2 relation types, 95 training triples and 10 evaluation and test triples.
The data sets are available on github \cite{caceres2021}.
Since the data sets are quite small, we report the performance on training, evaluation and test split here to guarantee a complete picture of the training process.

On both data sets, our model is capable of learning meaningful embeddings for nodes and relations in the graph, reaching similar performances as, e.g., TransE (\cref{tab3}).
Due to the event-based message passing, for the Brood War data set, only $(28.44 \pm 9.72) \%$ of aggregated spikes are used to produce the final spike embedding, and on the Federal States data set $(18.91 \pm 7.73) \%$ -- therefore being much sparser operations than in traditional, non-spiking GNNs that need all vector components of the aggregated embeddings for updates.
We are confident that the recent attention in optimizing simulating and training spiking neural networks \cite{mozafari2019spyketorch,wunderlich2020eventprop,perez2021sparse} will allow us to implement a considerably faster version of our proposed model in the near future that both speeds up the hyperparameter search to improve the presented results and allows us to apply the SR-GCN to larger KGs.

%% file: content/table2.tex
\begin{table*}[t]\renewcommand{\arraystretch}{1.2}
\caption{Test performances of SpikE and the hybrid model. Best values shown in bold.}\vspace{-3mm}
\begin{center}
\begin{tabular}{c | c c c| c c c}
\multicolumn{1}{c}{}&\multicolumn{3}{c}{UMLS}&\multicolumn{3}{c}{Countries S1} \\
\hline\hline
Model &  MRR& Hits@1& Hits@3 &MRR& Hits@1& Hits@3  \\
\hline\hline
SpikE \cite{dold2021spikeembed}   
 & \best{0.78} & 0.62 & \best{0.94} & 0.55 & 0.40 & 0.67 \\
w/ hinge loss & & & & & & \\
\hline
Hybrid 
  & 0.77 & \best{0.63} & 0.90 & \best{0.82} & \best{0.73} & \best{0.88}\\
 frozen + w/ self-loop & & & & & & \\
\hline\hline
\end{tabular}
\label{tab2}
\vspace{-3.5mm}
\end{center}
\end{table*}

%% file: content/table3.tex
\begin{table*}[!htb]\renewcommand{\arraystretch}{1.2}
\caption{Experimental results for TransE and SR-GCN. Best test values shown in bold.}\vspace{-4mm}
\begin{center}
\begin{tabular}{c | c | c c c| c c c}
\multicolumn{2}{c}{}&\multicolumn{3}{c}{StarCraft: Brood War}&\multicolumn{3}{c}{Federal States} \\
\hline\hline
Model & split & MRR& Hits@1& Hits@3 &MRR& Hits@1& Hits@3  \\
\hline\hline
TransE \cite{bordes2013translating}    & \tcol{train} & \tcol{1.00} & \tcol{1.00} & \tcol{1.00} & \tcol{0.71} & \tcol{0.42} & \tcol{1.00} \\
   & \tcolt{eval.} & \tcolt{0.70} & \tcolt{0.59} & \tcolt{0.77} & \tcolt{0.88} & \tcolt{0.80} & \tcolt{0.90} \\
   & test & \best{0.71} & \best{0.59} & \best{0.82} & \best{0.69} & \best{0.55} & \best{0.80} \\
\hline
SR-GCN & \tcol{train} & \tcol{0.99} & \tcol{0.98} & \tcol{1.00} & \tcol{0.71} & \tcol{0.44} & \tcol{0.99} \\
   frozen + w/ self-loop & \tcolt{eval.} & \tcolt{0.71} & \tcolt{0.63} & \tcolt{0.68} & \tcolt{0.85} & \tcolt{0.70} & \tcolt{1.00} \\
    & test & 0.67 & 0.50 & \best{0.82} & 0.56 & 0.35 & 0.70 \\
\hline\hline
\end{tabular}
\label{tab3}
\end{center}
\end{table*}

%% file: content/conclusion.tex
\section{Summary}
\label{sec:summary}

We propose a strategy to map R-GCNs for KG reasoning to an architecture that is closer to potential implementations on neuromorphic hardware, creating a first link between the fields of deep learning on graph-structured data and neuromorphic computing.
Our results address two challenges for mapping R-GCN-based models to neuromorphic architectures: (i) weight sharing introduced by the convolution operator, which requires non-local weight updates during learning and (ii) mapping the encoder-decoder structure of graph embedding models to spiking neural networks.
Concretely, we developed a model that composes a randomly initialized and frozen R-GCN encoder with a shallow decoder, which we subsequently mapped to spiking neurons.

In this context, we deploy a training strategy that does not train the weights of the R-GCN, but allows the gradient to flow through the network to tune the initial entity embeddings.
Since the aggregation of local neighborhood information is still intact with frozen weights -- basically acting as a form of graph-controlled routing -- the initial embeddings learn to utilize the frozen R-GCN to generate richer node embeddings.
By freezing the R-GCN weights, gradients and updates for much less parameters have to be calculated which greatly reduces the computational cost of our model compared to standard R-GCNs.  
We show this experimentally on an off-the-shelf processor using a standard PyTorch implementation, resulting in a significant speedup and a reduction of memory requirements during the backward pass -- between 20-90\% depending on the embedding dimension -- while keeping up or even outperforming other end-to-end, fully trained R-GCNs on the KBC task.
Even higher gains could be achieved by adding sparsity constraints to the filter weights \cite{sparsity2021}.

We further map the frozen R-GCN architecture to spiking neural networks, proposing a fully spike-based R-GCN model which extends previous work on spike-based graph embedding \cite{dold2021spikeembed} to inductive settings \cite{hamilton2017inductive} with dynamic graphs that can grow over time. 
Moreover, although not demonstrated in this work, the frozen R-GCN structure is sufficient to allow the application of state-of-the-art explainable techniques on graph data, which are most often based on masking the KG; changing the final embeddings to identify which parts of the graph are responsible for a certain link prediction or node classification outcome \cite{ying2019gnnexplainer,lucic2021cf}.
Apart from functional benefits, an intriguing property of SR-GCNs is the event-based calculation of embeddings in a first come first served fashion.
Conceptually, this purely temporal neighborhood aggregation strongly differs from how GNNs usually pool and combine embeddings, and leads to much sparser and potentially faster computation of neighborhood-aware embeddings.
For instance, in this work, we observe that only 20-30\% of the embeddings' vector components are used to update spike-based embeddings.
Although our current results are limited by the increased overhead of simulating and training spiking neurons (especially in the R-GCN layer), we are confident that more rigorous simulation code \cite{perez2021sparse} or emulations on accelerated neuromorphic devices \cite{göltz2020fast} will allow us to scale these models up to larger KGs and reach competitive performances.

\section{Conclusion}
\label{sec:conclusion}

Nowadays, KGs act as the backbone for various artificial intelligence tasks in numerous fields such as named entity disambiguation in NLP \cite{han2010structural}, visual relation detection \cite{baier2017improving} and visual question answering \cite{hildebrandt2020scene} in computer vision. Thereby, the underlying principle consists of condensing  structural  information in shallow KG embeddings which can subsequently be processed by other machine learning modules to perform various downstream tasks. Real-world, industrial applications that make use of this strategy are, for example, drug repurposing \cite{liu2021neural}, context-aware recommender systems \cite{hildebrandt2018configuration, hildebrandt2019recommender} and context-aware security monitoring \cite{soler2021graph}. 
We are convinced that our work constitutes an important step towards enabling a similar modular synthesis of neuromorphic machine learning methods and graph embedding algorithms, unlocking the potential of joining symbolic and numeric data to build powerful artificial intelligence applications and reasoning systems.
In particular, the proposed model can be used to learn spike-based representations of data structures that have no obvious or natural representation as spikes, e.g., social networks and tabular data, but can be modelled as a KG.

To the best of our knowledge, this constitutes the first deep learning architecture suitable for neuromorphic realizations that can reason on KGs, offering many attractive properties like sparse and resource efficient message passing between nodes in a KG.
Moreover, our results contribute to previous evidence \cite{jaeger2001echo,natschlager2002liquid,saxe2011random,raman2021frozen,yosinski2014transferable,isikdogan2020semifreddonets} that static connectivity motifs harbor potential functional benefits for neural networks without adding extensive computational costs -- especially when the surrounding neural structure is allowed to adapt to the static and frozen parts of the whole network -- ultimately reducing the amount of resources and complexity required for realization in application-specific integrated circuits.

%% file: content/acknowledge.tex
This work was partially funded by the Federal Ministry for Economic Affairs and Energy of Germany (BMWi) within the IIP-Ecosphere Project and by the German Federal Ministry for Education and Research (BMBF), funding project “MLWin” (grant 01IS18050).
We thank Serghei Mogoreanu and Josep Soler Garrido for helpful discussions and inspiration.
We further thank our colleagues at the Semantics and Reasoning Research Group and the Siemens AI Lab for their support.
\begin{figure}[h!]
 \begin{center}
    \includegraphics[width=0.18\textwidth]{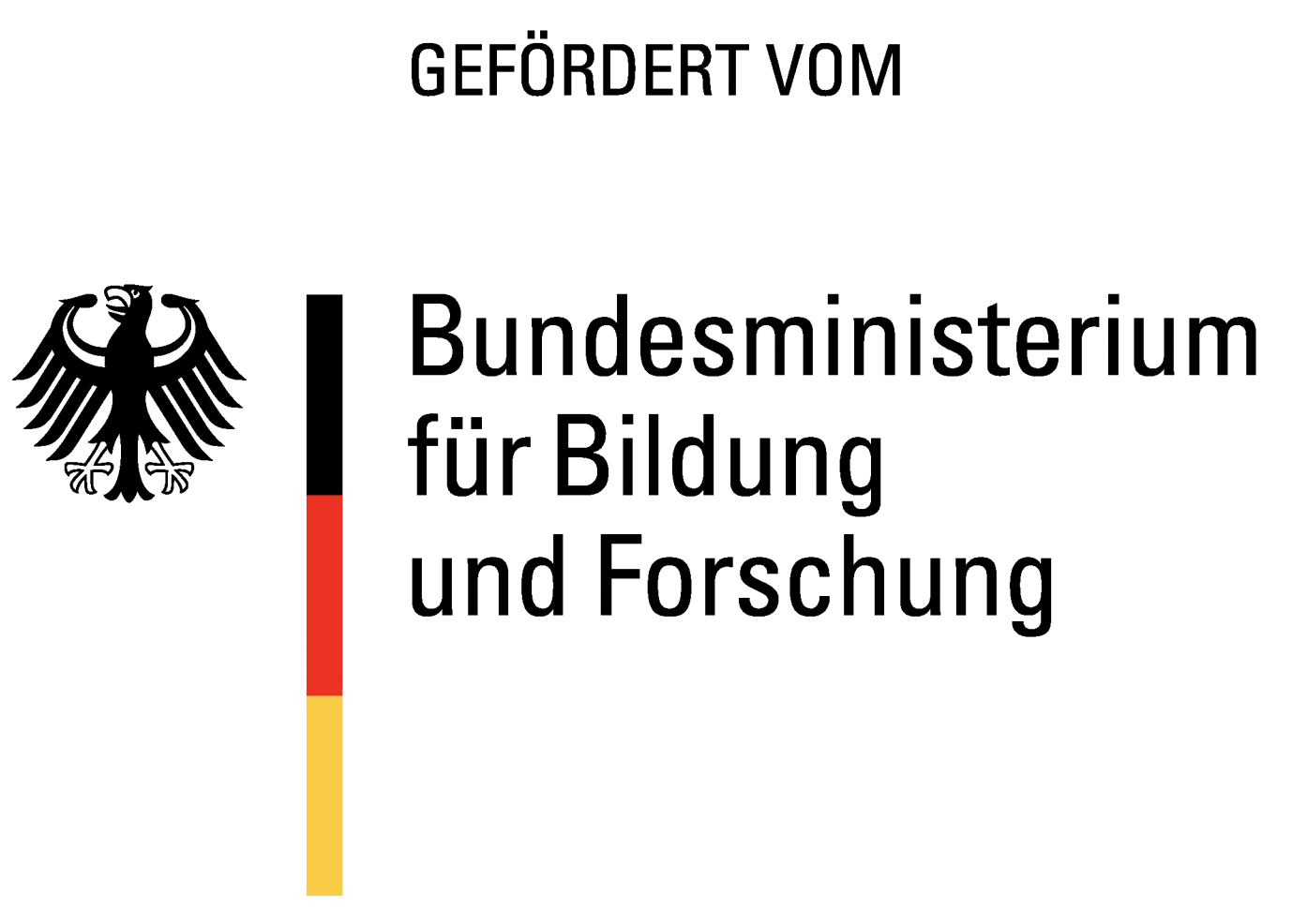}
 \end{center}
\end{figure}

%% file: content/details.tex
We based our implementations on the R-GCN code of the Deep Graph Library (DGL), see \cite{DGL}.

If not stated otherwise, we use the following parameters for all simulations: node embeddings are initialized randomly using a normal distribution $\mathcal{N}(0,1)$ and relation embeddings as well as R-GCN weights using Xavier initialization \cite{glorot2010understanding} with gain $\sqrt{2}$. 
For all non-spiking models, we use L2 regularization for node and relation embeddings with weighting $10^{-2}$.
We use 10 negative samples per training triple and use a default learning rate of $10^{-3}$.
Gradient updates are applied using the Adam optimizer \cite{kingma2014adam}.
The batch size is 64 except for FB15k-237, where we use a batch size of $2\cdot 10^4$.
Further, for FB15k-237 we sub-sample each batch by $50\%$ to calculate embeddings.
In all cases, we report the models that obtained the best performance on the evaluation split.
We always use only a single R-GCN layer without activation function, as we found that adding more layers does not improve the performance on the presented KBC tasks.

For all spiking models, we use $\taus = 0.5$, $\thresh = 1$ and an input spike time interval of $[-1, 1]$ (see \cite{dold2021spikeembed} for details).
For the initial spike embedding layer, we use weight regularization $\delta = 10^{-2}$ to ensure that all neurons spike, see e.g. \cite{dold2021spikeembed}, and the weights are initialized randomly from  $\mathcal{N}(0.2, 1)$.

For the results reported in \cref{tab1}, embedding dimensions were obtained by running experiments with dimensions of \{16, 32, 64, 128\}.
For TransE, the best embedding dimension is \{32, 64, 128\}, for the R-GCN + TransE \{128, 128, 128\}, for the R-GCN + TransE with frozen weights \{128, 128, 64\}, for the R-GCN + TransE with self-loop \{64, 128, 128\} and for the R-GCN + TransE with frozen weights and self-loop \{64, 128, 128\} -- with entries corresponding to \{Countries, UMLS, FB15k-237\}.
For DistMult, we use \{6, 32, 32\} with learning rates \{$5\cdot 10^{-2}$, $10^{-3}$, $10^{-3}$\} and for the R-GCN + DistMult \{56, 32, 128\} with learning rates \{$10^{-3}$, $10^{-2}$, $10^{-3}$\}.
For all R-GCN models, we use dropout of $0.2$ during training.

For the results reported in \cref{tab2}, we use the following \{dimension, number of input neurons, learning rate\}:
SpikE: \{32, 40, $10^{-2}$\} for Countries and \{32, 20, $10^{-2}$\} for UMLS; Hybrid: \{64, 40, $10^{-3}$\} for both data sets.

For the results reported in \cref{tab3}, we initialize the frozen SR-GCN weights randomly from $\mathcal{N}(1,5)$.
For the Federal States data set, we further use only 5 negative examples per training triple.
No dropout is used in these experiments.
For all TransE experiments, we use an embedding dimension of 16, except for TransE without R-GCN layer on the Brood War data set where we use 32.
For the spike-based convolutions, we use \{dimension, number of input neurons\} of \{16, 16\} for Brood War and \{32, 16\} for Federal States.